\pdfoutput=1

\documentclass[journal]{IEEEtran}
\usepackage[ruled,linesnumbered]{algorithm2e}

\usepackage{bbding}

\usepackage{times}
\usepackage{epsfig}
\usepackage{graphicx}
\usepackage{subfigure}

\usepackage{amsfonts}
\usepackage{underscore}
\usepackage{pifont}
\usepackage{color}
\usepackage[table]{xcolor}

\definecolor{bg}{RGB}{225,236,247}

\usepackage{xcolor}
\usepackage[normalem]{ulem} 
\newcommand \hl{\bgroup\markoverwith
  {\textcolor{bg}{\rule[-.5ex]{2pt}{2.5ex}}}\ULon}

\usepackage{amsthm,amsmath,amssymb}       
\usepackage{mathrsfs}

\usepackage{makecell}
\usepackage{multirow}
\usepackage{hyperref}
\hypersetup{nolinks=true}

\ifCLASSINFOpdf
\else
\fi
\begin{document}
%
\title{Weakly-supervised Part-Attention and Mentored Networks for Vehicle Re-Identification}
%
%
%

\author{Lisha Tang,~Yi Wang,~\IEEEmembership{Member,~IEEE,}
        Lap-Pui Chau,~\IEEEmembership{Fellow,~IEEE}
\thanks{Lisha Tang and Yi Wang are with School of Electrical and Electronic Engineering, Nanyang Technological University, Singapore, 639798 (e-mail: lisha001@e.ntu.edu.sg, wang1241@e.ntu.edu.sg).}
\thanks{Lap-Pui Chau was with the School of Electrical and Electronic Engineering, Nanyang Technological University. He is now with the Department of Electronic and Information Engineering, The Hong Kong Polytechnic University. (e-mail: lap-pui.chau@polyu.edu.hk).}
\thanks{Corresponding author: Yi Wang.}}

%

%

\markboth{Journal of \LaTeX\ Class Files}%
{Shell \MakeLowercase{\textit{et al.}}: Bare Demo of IEEEtran.cls for IEEE Journals}
%



\maketitle
\begin{abstract}
Vehicle re-identification (Re-ID) aims to retrieve images with the same vehicle ID across different cameras. Current part-level feature learning methods typically detect vehicle parts via uniform division, outside tools, or attention modeling. However, such part features often require expensive additional annotations and cause sub-optimal performance in case of unreliable part mask predictions. 
In this paper, we propose a weakly-supervised Part-Attention Network (PANet) and Part-Mentored Network (PMNet) for Vehicle Re-ID. Firstly, PANet localizes vehicle parts via part-relevant channel recalibration and cluster-based mask generation without vehicle part supervisory information. Secondly, PMNet leverages teacher-student guided learning to distill vehicle part-specific features from PANet and performs multi-scale global-part feature extraction. During inference, PMNet can adaptively extract discriminative part features without part localization by PANet, preventing unstable part mask predictions. We address this Re-ID issue as a multi-task problem and adopt Homoscedastic Uncertainty to learn the optimal weighing of ID losses. Experiments are conducted on two public benchmarks, showing that our approach outperforms recent methods, which require no extra annotations by an average increase of 3.0\% in CMC@5 on VehicleID and over 1.4\% in mAP on VeRi776. 
Moreover, our method can extend to the occluded vehicle Re-ID task and exhibits good generalization ability.

\end{abstract}

\begin{IEEEkeywords}
Vehicle re-identification, weak supervision, attention, multi-task learning.
\end{IEEEkeywords}

%
\IEEEpeerreviewmaketitle

\begin{figure}
  \centering
  \subfigure[]{
    \label{fig:subfig-a} 
    \includegraphics[scale=0.40]{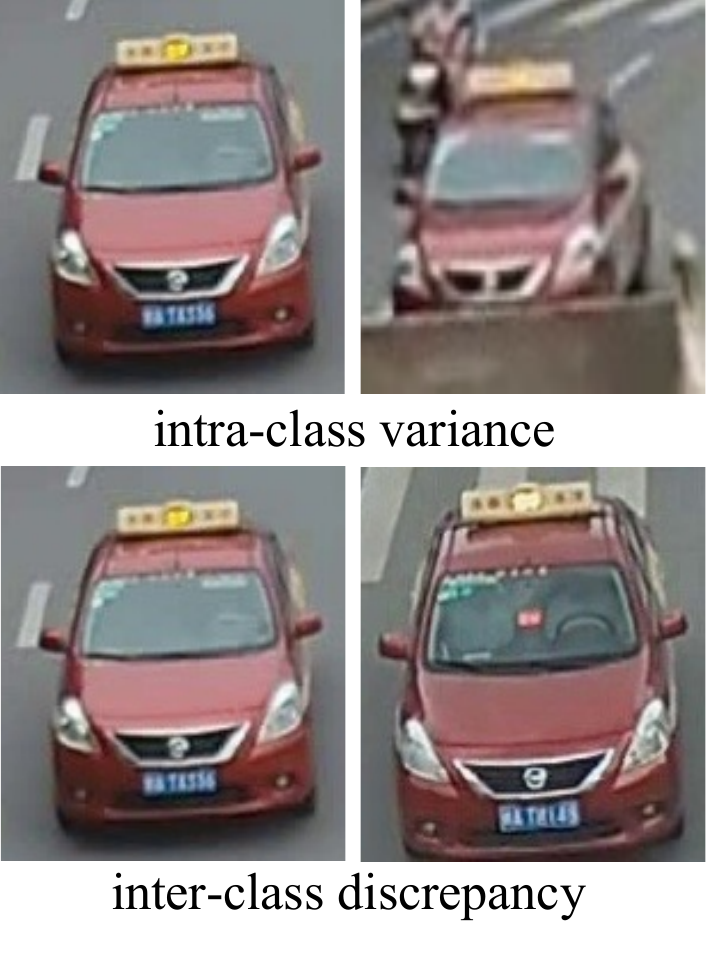}
    }\quad
  \subfigure[]{
    \label{fig:subfig-b} 
    \includegraphics[scale=0.40]{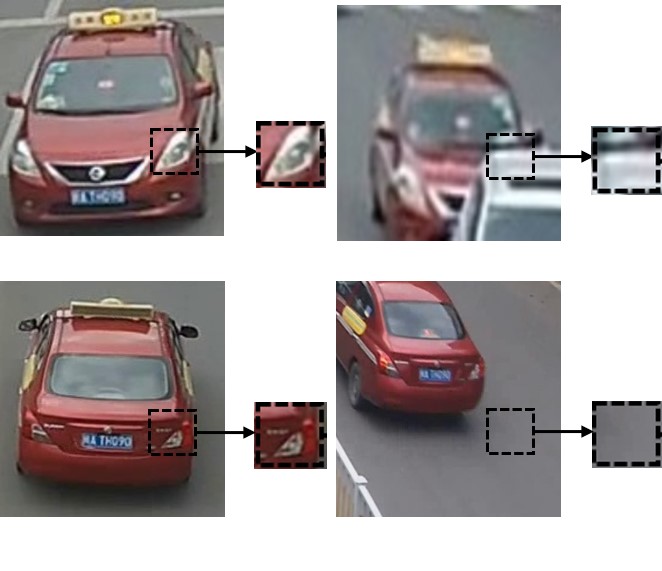}
    }
  \subfigure[]{
    \label{fig:subfig-c} 
    \includegraphics[scale=0.63]{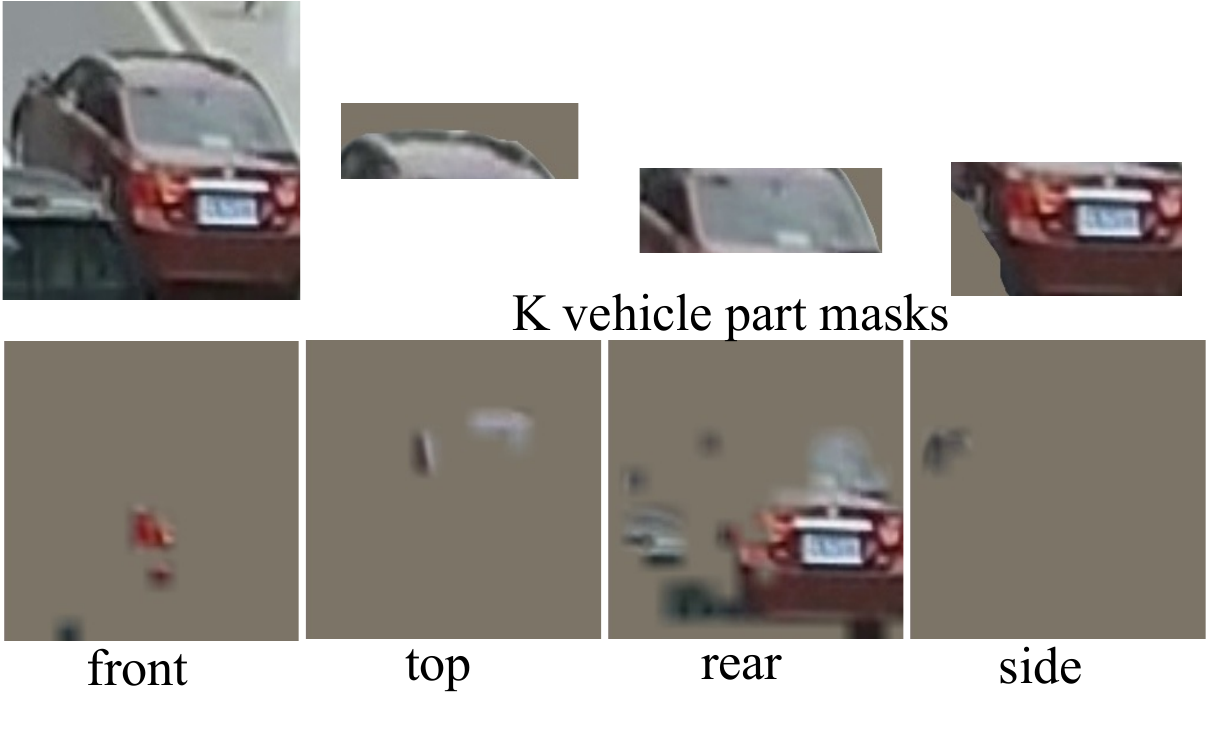}
    }
  \caption{(a) Two images in the first row show the large intra-class variance of the same vehicle caused by complex variations in illumination, image quality, viewpoint, and background clutter, whilst the images in the second row illustrate the minor inter-class discrepancy caused by near-duplicated vehicles. (b) Four zoomed-in patches show that the same spatial position across images may correspond to different vehicle parts due to occlusion, viewpoint changes, and diverse spatial distributions. (c) Images on the first row represent the sample image and $K=3$ part masks by our PANet, and figures on the second row show four view-aware masks generated by PVEN \cite{r11}.}
  \label{fig1} 
\end{figure}

\section{Introduction}
\label{sec-Intro}
\IEEEPARstart{G}{iven} a query vehicle image, vehicle Re-ID aims at identifying the images with the same identity in a gallery set across multiple non-overlapping cameras. It holds great potential in public security and intelligent transportation, thereby drawing increasing attention from academia and industry. However, this task still suffers from the large intra-ID variance and the subtle inter-ID discrepancy (see Fig.\ref{fig:subfig-a}) due to complex variations in illumination, poor image quality, background clutter, and viewpoint. These factors result in spatial component misalignment as shown in Fig.\ref{fig:subfig-b}, where the same spatial positions across two images may correspond to different vehicle parts.

To address this issue, recent works which elaborately learn discriminative vehicle part-level representations were proposed. Typically, there are three categories. The first category \cite{r8,r12} uniformly partitions an image into patches or horizontal stripes to learn part-level features. However, such naive uniform division may not be well aligned with vehicle parts owing to viewpoint changes, illumination, occlusion, and diverse spatial distributions in the image. The second category parses the image into semantic-aware or view-aware parts by using outside tools \cite{r11,r9,r26,r10,r22,r16,r53}. For example, He \emph{et al.} \cite{r9} pre-defined windows, lights, and brands for each vehicle and used them to train a YOLO \cite{r20} detector. PVEN \cite{r11} utilizes a U-Net model to parse each image into four views based on the 20 key-points labeled. Nevertheless, these methods require substantial pixel-level annotations, which are time-intensive and error-prone. Additionally, since the masks predicted by the outside tools may not be sufficiently reliable, it is sub-optimal to directly rely on them in vehicle Re-ID task (see the second row of Fig.\ref{fig:subfig-c}). The third type learns fine-grained part-level features by incorporating attention modeling \cite{rattn1,rattn2}. Although attention-based methods tend to be more efficient than outside tool-based ones, they still face challenges, such as background clutter. It is also notable that the training process for some attention-based methods \cite{rattn1} is quite complex.

\begin{figure*}[t]
\centering
\includegraphics[scale=0.53]{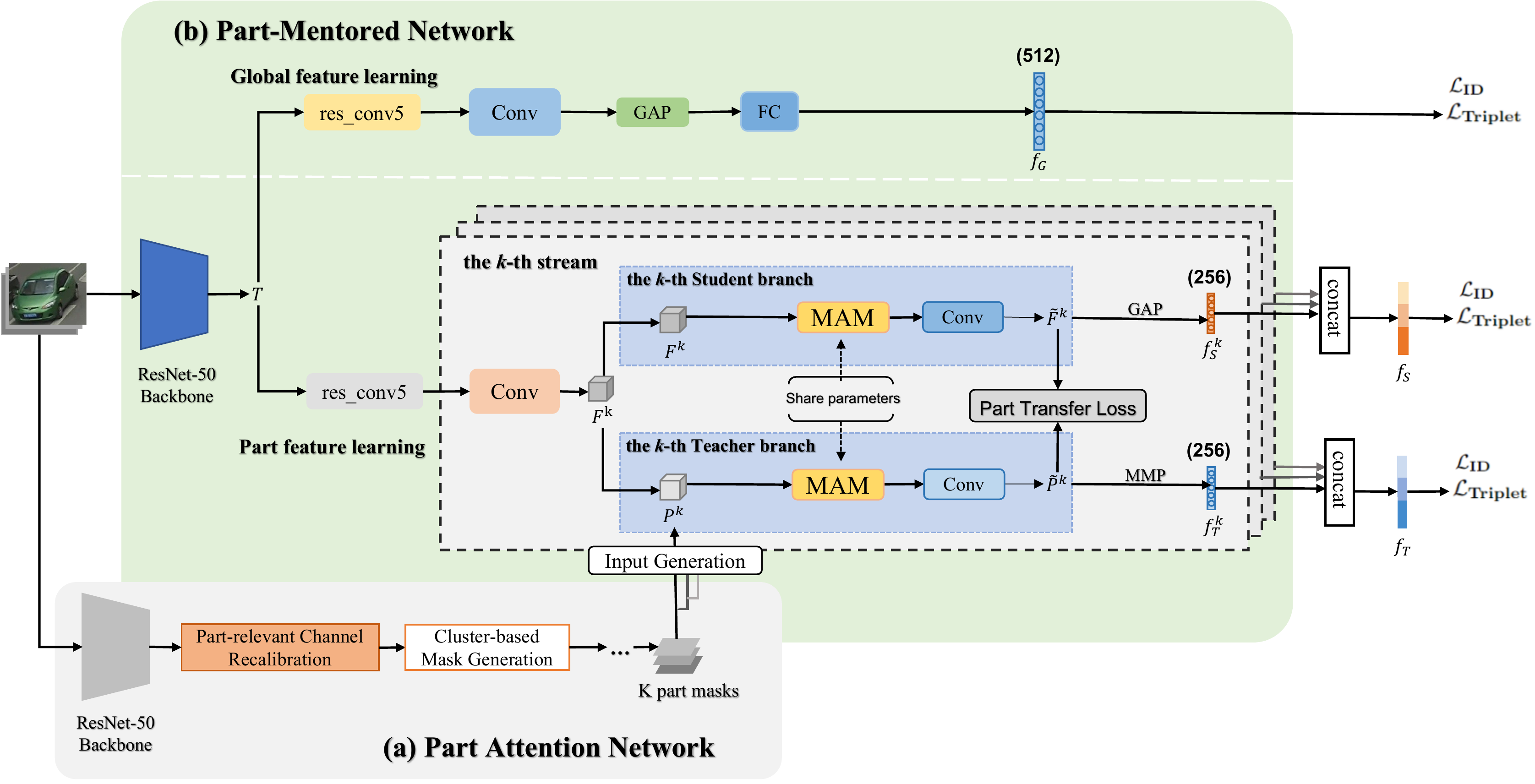}
\caption{Architecture of the proposed Part Attention Network (PANet) and Part-Mentored Network (PMNet). (a) PANet performs $K$ vehicle part localization via part-relevant channel recalibration and cluster-based mark generation module. (b) To learn robust part-level features, PMNet applies the part masks from PANet to $K$ streams, each consisting of a Student branch and a Teacher branch. The part-level knowledge learnt by each Teacher branch is transferred to the correspondent Student branch, such that the Student branch can be retained during inference free from prior part masks by PANet. Then, the part feature learning branch is cooperated with the global feature learning branch to implement the global-part feature extraction. Multi-task learning is applied to train PMNet. For simplicity, ReLU and BatchNorm layer following each Conv layer is omitted by default.}
\label{Fig:PGLN}
\end{figure*}

Accordingly, we propose a novel Part Attention Network (PANet) for vehicle part localization under weak supervision and a Part-Mentored Network (PMNet) for part-specific feature extraction and global-part feature aggregation, as shown in Fig.\ref{Fig:PGLN}. To address spatial misalignment without additional annotations, PANet predicts a refined foreground mask and robustly locates different prominent vehicle parts based on part-relevant channel recalibration and cluster-based mask generation (see the first row of Fig.\ref{fig:subfig-c}). Afterward, PMNet performs global and part-level feature learning. Specifically, in the part-level feature learning branch, PMNet applies teacher-student guided learning to distill high-quality part-specific representation. During inference, PMNet's students can independently extract part features without teachers, which may be affected by inaccurate part masks of PANet. Briefly, our contributions are four-fold:

\begin{enumerate}
\item We introduce PANet to locate different vehicle components in an attention manner, which is easy to optimize. This attempt leads to more accurate part mask predictions, reducing the influence of background interference and freeing from extra vehicle part supervision.
\item We propose a teacher-student guided learning structure in PMNet to transfer learnt vehicle part knowledge from teachers to students and learn more robust part-level representations. During inference, with strong students, PMNet can work independently without part mask predictions from PANet.
\item An end-to-end multi-task learning scheme is introduced to learn discriminative global features and vehicle part features of students and teachers. The combined features boost the model generalization, even for occluded vehicle images. 
\item Extensive experiments and ablation studies are conducted to demonstrate the superiority of our method compared with recent approaches. Experiments on occluded vehicle test sets also show that our method effectively resists the occluded vehicles.
\end{enumerate}

The remainder of the paper is organized as follows. Related works are reviewed in Section \ref{SecRelatedWork} and the detailed model structure is described in Section \ref{sec3}. Afterwards, Section \ref{SecEXP} presents implementation details and experiments, followed by the conclusion in Section \ref{SecConclusion}.

\section{Related Work}
\label{SecRelatedWork}
\subsection{Vehicle Re-ID}
Derived from person Re-ID, vehicle Re-ID has attracted increasing attention, but the performance is far from satisfactory. Robust, high-quality feature representations are key to this task. As one of the pioneer works to study vision-based vehicle re-ID model, Ref. \cite{r27} combines hand-crafted features like SIFT and Color Name with deep features extracted from CNN models. With recent breakthroughs in deep learning, experiments demonstrate that deep features are more discriminative than hand-crafted ones. Since then, deep learning-based approaches have dominated vehicle Re-ID.

\subsubsection{Attribute-based feature learning}

Early deep learning methods directly incorporated various meta information such as vehicle attributes (e.g., model, color) and spatial-temporal information to enhance global representations \cite{r28,r71,r73}. Ref. \cite{r28} fuses camera views, vehicle types, and color into feature embedding. Ref. \cite{r71} proposes a coarse-to-fine ranking loss with vehicle model information to guarantee both intra-class compactness and inter-class distinction. However, these holistic image-based approaches typically suffer from over-fitting problems and miss detailed local clues, which are crucial to differentiate similar samples. Besides, attributes and spatial-temporal cues are not always available, limiting the application and deployment of those models. To resolve this, regional feature learning is introduced for more discriminative, powerful representations.

\subsubsection{Regional feature learning}

Current regional feature learning methods can be grouped into three categories. The first category \cite{r12,r8,r58} utilizes uniform division, which divides the image or feature map into patches or stripes. However, as is mentioned in the Introduction, such naive division suffers from part misalignment due to complex variations in illumination, occlusion, and spatial distributions. The second strategy aims to learn viewpoint-aware or semantic-aware features via outside tools \cite{r11,r9,r26,r10,r22,r53}. For instance, SPAN \cite{r10} predicts masks for different views of a vehicle and emphasizes co-occurrence parts when computing the feature distances. Zhao \emph{et al.} \cite{r22} collected a vehicle dataset with 21 classes of attribute labels and trained a Single-Shot Detector \cite{r25}. Nevertheless, this sub-category requires substantial human annotations, and the results produced by the outside tools may not always be reliable, thus leading to performance degradation in Re-ID. 

Another type models attention selection in vehicle part detection and Re-ID \cite{r52,r26,rattn1,rattn2,r72}. For example, MVAN \cite{r52} proposes a multi-view branch network, where each branch learns a viewpoint-specific feature with a global and local attention. However, this method generally applies attention weighting to the entire image box and fails to compare the subtle yet discriminative details within each vehicle part. Instead, \cite{r26,rattn1,rattn2} models both hard and soft attention selection to perform coarse-to-fine identity searching. For example, TAMR \cite{r26} leverages two STNs \cite{r30} as hard attention to locate windscreens and car heads and employs an adapted Residual Attention Module as soft attention to achieve pixel-wise refinement. However, this approach requires additional data annotations about positions of windscreens and car-head parts, while information within other informative regions (i.e., front bumper, tires) is ignored in this method. HA-CNN \cite{rattn2} jointly units multiple complementary attention but still suffers from unreliable part localization results in case of severe misalignment or occlusion. Therefore, based on a two-stage attention combination, our method explores weakly-supervised vehicle part localization in a simple architecture and extracts multi-scale details during the second-stage attention refinement. Additionally, we transfer part concepts in a teacher-student manner, which frees our PMNet from prior part mask prediction during inference.

\subsubsection{Transfer learning}
Transfer learning is recently introduced to person Re-ID \cite{r34,r35,r38,r79,r80}. For instance, a DSAG-Stream designed in \cite{r34} acts as a regulator to guide an MF-Stream for densely semantically aligned feature learning. Ref. \cite{r38} handles body part misalignment with prior part-aware supervision in a transfer learning manner. However, Ref. \cite{r38} relies on outside tools during body part detection, which might not always be reliable, especially in case of low image quality or background interference. Refs. \cite{r79,r80} performs horizontal refinement on Uniform Division, which might show limited ability in handling part alignment, especially for vehicles. Relevant exploration in vehicle Re-ID is quite insufficient. Inspired by \cite{r38}, our PMNet transfer part concept learnt to holistic image-based branches, mitigating occlusion and reducing the computational cost required for vehicle part localization during inference.

\subsubsection{Occluded Re-ID}
Different from traditional Re-ID task, which performs pedestrian or vehicle retrieval in the full-body domain, occluded Re-ID is a challenging practical issue. So far, occluded person Re-ID has attracted much attention, but there is still little research related to occluded vehicle Re-ID. For example, Ref. \cite{r56} boosts the model generalization by synthesizing occluded samples. ASAN \cite{r55} proposes a CAM-based segmentation module and a shift feature adaptation module to learn features within the visible region of the image. It designs the first occluded test set based on VeRi776 \cite{r27} but requires extra attribute information, which is not always available.

\subsection{Weakly-supervised localization}

Refs. \cite{r51,r47,r57} also perform weakly-supervised localization based on the principle that different channel groups tend to exhibit strong activation on different semantic regions. For instance, PAN \cite{r57} re-locates the pedestrian with STN \cite{r30} and merges it with the global feature to suppress background interference. However, this method merely concentrates on re-localization of the full human body but fails to take advantage of the fine-grained details which can be provided by different body parts. PL-Net \cite{r47} directly clusters the coordinates with the maximum response of each channel for component localization, which is proved quite unstable because the encoded feature maps are sensitive to noises and background clutter. To resolve these, our proposed method mitigates noises and background interference via part-relevant channel recalibration and clustering, and foreground mask refinement.


\section{Methodology}
\label{sec3}
As is illustrated in Fig.\ref{Fig:PGLN}, our proposed method consists of two networks: Part-Attention Network (PANet) and Part-Mentored Network (PMNet). Without extra vehicle part supervision, PANet is proposed to predict vehicle part masks via part-relevant channel recalibration and cluster-based mask generation. Then, the masks are applied to PMNet to learn robust part features in a teacher-student manner by the part transfer loss. During inference, only the student branch can be retained, free from the part masks by PANet. They will be detailed in Sec. \ref{sec3A} and Sec. \ref{sec3B}, respectively. The overall multi-task learning scheme is introduced in Sec.\ref{sec3C}.

\begin{figure}[t]
\centering
\includegraphics[height=180pt,width=255pt]{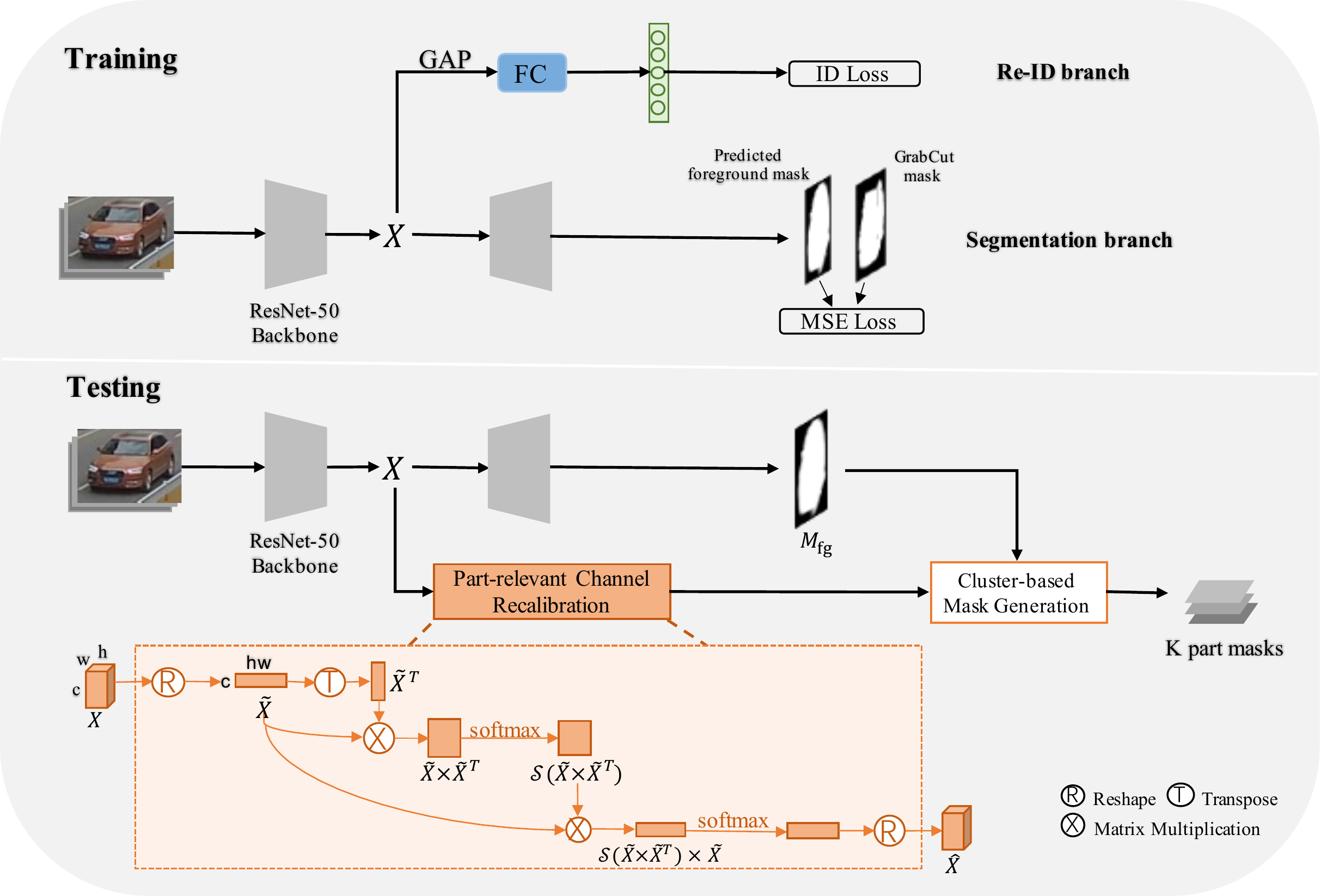}
\caption{Architecture of Part Attention Network (PANet). As the upper sub-figure illustrates, PANet consists of the ResNet-50 backbone, classification branch for Re-ID \cite{r43}, and segmentation branch. PANet is supervised by an ID loss and a Mean Square Error loss during training. The BatchNorm applied to the fully connected (FC) layer is omitted for simplicity. The lower sub-figure shows the pipeline to predict $K$ dense vehicle part masks during testing, and details of Part-relevant Channel Recalibration module are exhibited in the orange rectangle. $\mathcal{S}(.)$ is the Softmax normalization.}
\label{fig2}
\end{figure}

\subsection{Part Attention Network for Vehicle Part Localization}
\label{sec3A}
The goal of PANet is to predict $K$ vehicle part masks under weak supervision. Recent works have shown that different channel groups in the top layers of CNNs describe different body parts in person Re-ID \cite{r47}. Similar correspondences between channels and vehicle parts are also observed in vehicle Re-ID. Based on this, one intuitive solution is to spatially locate different prominent parts by exploring such inter-channel relationships. Nevertheless, results from \cite{r47} are quite unstable and sensitive to background interference as well as illumination. To learn more robust part features, the proposed PANet performs part-related channel recalibration via attention mechanism and produces a refined foreground mask to suppress influences of cluttered background.

As depicted in Fig.\ref{fig2}, based on ResNet-50 backbone, PANet encodes the $i$-th image into the encoded feature map, $X_{i}$. We omit the index $i$ for simplification. In the decoder, PANet combines a classic identity classification branch and a segmentation branch during training. The classification branch extracts instance-specific features to weigh different vehicle parts, while the segmentation branch predicts the foreground masks. In the classification branch, the encoded feature map $X$ is fed into a Global Average Pooling (GAP) and a fully-connected (FC) layer to compute ID loss. In the segmentation branch, $X$ goes through a decoder which comprises four transposed convolutional layers and a Sigmoid. By default, the first three transposed convolutional (Conv) layers are followed by a BatchNorm and ReLU, respectively. To obtain foreground mask labels for segmentation without any manual annotations, inspired by \cite{r10}, we generate a coarse foreground mask of each vehicle by using the simple handcrafted algorithm, GrabCut \cite{r40}. These coarse masks are error-prone and thus serve as pseudo labels to supervise the segmentation process with a Mean Square Error (MSE) loss.

During testing, PANet produces a refined foreground mask from the segmentation branch. Meanwhile, the encoded $X$ is processed by a Part-relevant Channel Recalibration (PCR) module and a Cluster-based Mask Generation (CMG) module to predict $K$ vehicle part locations. By cropping the foreground mask according to these coordinates, $K$ dense part masks are obtained as input for PMNet. We describe the two key components in detail, i.e., PCR and CMG.

\subsubsection{Part-relevant Channel Recalibration} 
As is mentioned above, the encoded feature map $X$ is unstable, and utilizing it for part mask generation often leads to suboptimal results \cite{r47}. To convert $X$ into a more robust and consistent attention map, PCR module applies the attention mechanism on $X$, which adaptively highlights the relevant channels for each respective vehicle part and suppresses unrelated information, as shown by the expanded orange block in Fig.\ref{fig2}. Given feature $X \in \mathbb{R}^{C\times H\times W}$ with the channel number of $C$, the height of $H$, and width of $W$, we reshape this feature map into a matrix, $\widetilde{X} \in \mathbb{R}^{C\times HW}$. Then, we compute the similarity between $C$ channels, i.e.,

\begin{equation}
\begin{aligned}
G(\widetilde{X})=\mathcal{S}\left(\widetilde{X}\times \widetilde{X}^{T}\right),
\end{aligned}
\end{equation}
where $\mathcal{S(.)}$ is the Softmax normalization over the second dimension of the matrix. The larger value of $G(\widetilde{X})_{jk}$ is, the more similar channel $j$ and channel $k$ are; namely, the more likely channel $j$ and channel $k$ concentrate on the same vehicle part. Subsequently, information within the channels of $\widetilde{X}$ is recalibrated by $G(\widetilde{X}) \times \widetilde{X}$. The normalized attention map, $\hat{X}$, is obtained by
\begin{equation}
\begin{aligned}
\hat{X}=\mathcal{S}\left(G(\widetilde{X}) \times \widetilde{X}\right)=\mathcal{S}\left( \mathcal{S}\left(\widetilde{X}\times \widetilde{X}^{T}\right)\times \widetilde{X}\right),
\end{aligned}
\end{equation}
Finally, the attention map is reshaped back to the shape of $X$ as the output of PCR. 

\IncMargin{1em}
\begin{algorithm}[t]
\caption{Part Mask Generation}\label{algorithm}
  \SetKwData{Left}{left}\SetKwData{This}{this}\SetKwData{Up}{up}
  \SetKwFunction{Union}{Union}\SetKwFunction{FindCompress}{FindCompress}
  \SetKwInOut{Input}{input}\SetKwInOut{Output}{output}

  \Input{The attention map $\hat X \in \mathbb{R}^{C\times H\times W}$ and the refined foreground mask $M_{fg}$}
  \Output{$K$ vehicle part masks}
  \BlankLine
  \emph{First phase - $K$ part masks clustering}:
  
  \For{$c\leftarrow 1$ \KwTo $C$}{
    Obtain the bounding box of the largest connected component (morphology analysis) for $\hat X_c \in \mathbb{R}^{1 \times H\times W} \leftarrow BBox(\hat X_c)$;
    
    Compute the central coordinate $(x_c,y_c)$ of $BBox(\hat X_c)$.}
    Conduct k-means clustering on the central coordinate set $\{(x_c,y_c)\}_{c=1}^{C}$ and get $K$ clusters $\mathcal{C}=\left\{{\mathcal{C}}_{k}\right\}_{k=1}^{K}$.
    
    \emph{Second phase - random selection of part marks during each training iteration}:
    
    \For{$k\leftarrow1$ \KwTo $K$}{\label{forins}
    Randomly select a central coordinate $(x_{r},y_{r})$ in cluster ${\mathcal{C}}_{k}$
    
    Obtain $BBox(\hat X_r)$
    
    Crop $BBox(\hat X_r)$ according to $M_{fg}$ and get the $k$-th part mask.}
\end{algorithm}\DecMargin{1em}

\subsubsection{Part Mask Generation}
The generation of part masks is demonstrated in Algorithm \ref{algorithm}. Based on morphology, we first find the largest connected component for each channel of the feature map, $\hat X_c \in \mathbb{R}^{1 \times H\times W}, c \in \{1,2,...,C\}$, and obtain the corresponding bounding boxes $BBox(\hat X_c)$. The connected component represents the significant response of vehicle part features, resistant to illumination variations or background clutter. Then, the central coordinates of $BBox(\hat X_c)$ are grouped into $K$ clusters by the k-means clustering. Since there is no supervisory information to determine the boundaries of $K$ semantic parts, instead of pre-defining each part mask's size, we convert this step into a probability model \cite{r70}, in which we randomly select one bounding box from each cluster and use the box to crop the refined foreground mask at each training iteration of PMNet. The random sampling manner provides a glance of vehicle part features for the Teacher branches of PMNet and guides their part attention (see Sec. \ref{sec3B}).

\textbf{Remark.} Although both our method and SPAN \cite{r10} employ GrabCut for pseudo label generation, our PANet performs vehicle part localization through attention without extra labels, while SPAN generates viewpoint-aware masks trained by viewpoint labels. With the guide of PANet, our PMNet beats SPAN with a remarkable increase of over 10\% in mAP on both VeRi776 and VehicleID (see Sec. \ref{SecExpVeri776}).



\subsection{Part-Mentored Network for Vehicle Re-ID}
\label{sec3B}
Given $M$ training images $\mathcal{I}=\left\{\boldsymbol{I}_{i}\right\}_{i=1}^{M}$ with the corresponding identity labels as $\mathcal{Y}=\left\{{y}_{i}\right\}_{i=1}^{M}$ captured by multiple non-overlapping cameras, we aim to learn a multi-grained high-quality deep feature representation for vehicle Re-ID in case of complex illumination variations, image quality, and occlusion. Existing works usually learn features by separating this task into two parts: part detection and global-part feature aggregation \cite{r11,r9,r26,r10,r22}, which poses two downsides. Firstly, their prior part detection by outside tools or attention-based detectors during both training and testing requires extra computational cost and additional manual annotations. Secondly, these detection results are not always reliable, especially for low-quality images, thus leading to suboptimal performance. To this end, we formulate a Part-Mentored Network (PMNet), which drives to learn part-specific features in a teacher-student manner and maximizes complementary information with multi-scale attention. Since the first step, PANet, does not require additional annotations, our entire method is weakly-supervised in the context of optimizing Re-ID performance. Besides, our PMNet can retain Student branches and bypass part detection (i.e., the output of PANet) during testing, preventing from unstable detection results and saving computational cost.

As is shown in Fig.\ref{Fig:PGLN}, PMNet consists of a ResNet-50 backbone and two heads. (1) Global feature learning head: this aims to learn the optimal holistic features from the entire vehicle images. (2) Local (Part) feature learning head composed of $K$ streams with the identical structure: each stream aims to learn the most discriminative visual features for one vehicle part localized by the first step, PANet. In specific, PMNet adopts two \emph{res_conv5} residual stages to separate the two heads. 
We remove the last spatial down-sampling operation of both \emph{res_conv5} against spatial information loss.

\begin{figure}[t]
\centering
\includegraphics[scale=0.67]{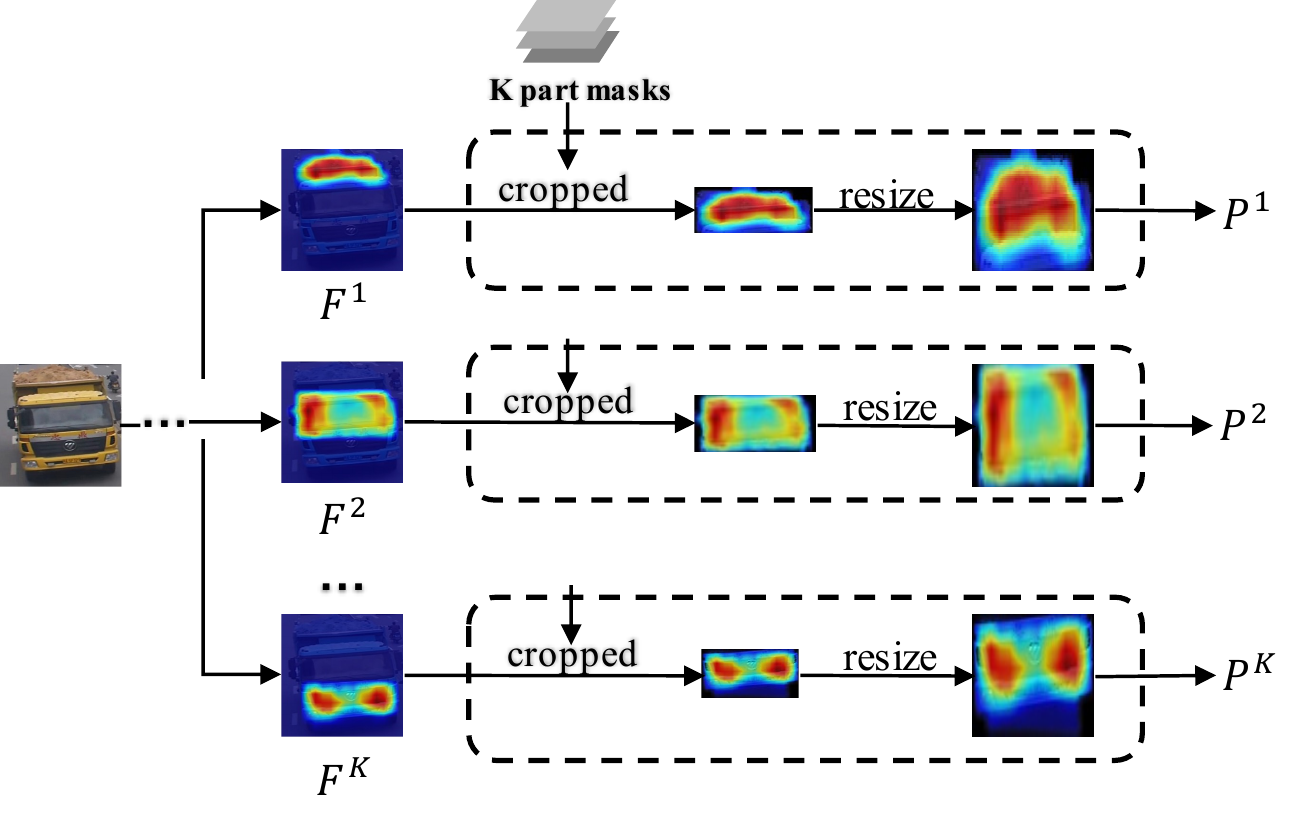}
\caption{Input generation for Teacher branches. The dashed block shows details of the Input Generation module in Fig.\ref{Fig:PGLN}. Here we set $K=3$ in our experiment and show visualization maps of the three inputs for three Teacher branches. These maps respectively focus on three different vehicle parts with semantic meanings, i.e., vehicle roof, windscreen, and headlights.}
\label{figPT}
\end{figure}

\subsubsection{Global feature learning} 
Our global head feeds the output feature map of the \emph{res_conv5} into a $1 \times 1$ Conv layer, a GAP, a FC layer, and a classification layer \cite{r43} sequentially, obtaining the global feature, $f_G$. The training process is supervised by one Triplet loss \cite{r46} and one ID loss on $f_G$.

\subsubsection{Part feature learning} 
Our local branch comprises $K$ streams, which respectively concentrate on $K$ different vehicle parts. Each stream firstly utilizes one $1\times1$ Conv layer to obtain the shared features ($F^k, k\in \{ 1,2,..,K\}$). Then, we propose two learning branches, i.e., Student and Teacher branches, which respectively focus on the holistic features and the specific vehicle part features. The Students learn vehicle part features from entire feature maps with the guidance of Teachers trained on part-cropped feature maps. Compared with existing works \cite{r9,r10,r11,r12,r16} that adopt the single local branch (Teacher only) for one part mask, we do not totally trust part masks input to Teachers (denoted as noisy Teacher). Instead, we build the Students to learn more robust part features from a holistic view. 

All the Students and Teachers are similar in architecture but have different inputs. Specifically, each Student feeds $F^k$ into one Multi-scale Attention Module (MAM), one Conv layer, and one GAP successively. The GAP learns translation- and scale-invariant holistic feature. Each Teacher has the same structure as the Student except for replacing GAP with a Mask Max Pooling (MMP) layer. MMP can be implemented as a Max Pooling layer with pooling size equal to the area of part masks. Compared with GAP, MMP helps mine significant local features within the constrained part mask region. The major difference between Teacher and Student branches lies in their inputs. The input of all Students is the original feature map $F^k$, while the input of the Teacher is $P^k$ where $F^k$ is cropped by the $k$-th vehicle part mask of PANet. Fig.\ref{figPT} illustrates the inputs of Teacher branches, in which the feature maps ($F^k$) are separately cropped according to the part masks. To be consistent with Students' input size, the cropped feature map is resized to the shape of $F^k$, generating $P^k$. 

During training, the Teachers guide their corresponding Students to concentrate on the specific vehicle parts by the Part Transfer loss. The Part Transfer loss will be presented in detail in Sec.\ref{SecPartTransfer}. Since Students learn features within the entire feature map, $F^k$, it avoids degrading by the noises from inaccurate vehicle part masks. Moreover, these Teacher branches can be removed during inference, saving the computation cost required for prior part masks by PANet. Our method achieves state-of-the-art performance without Teachers, i.e., our PMNet can be free from prior part masks of PANet (Refer to Sec. \ref{SecExpVeri776}). 
Next, we will introduce the details of Multi-Scale Attention, batch-wise Part Transfer loss, and Re-ID loss functions in PMNet.

\begin{figure}[t]
\centering
\includegraphics[scale=0.48]{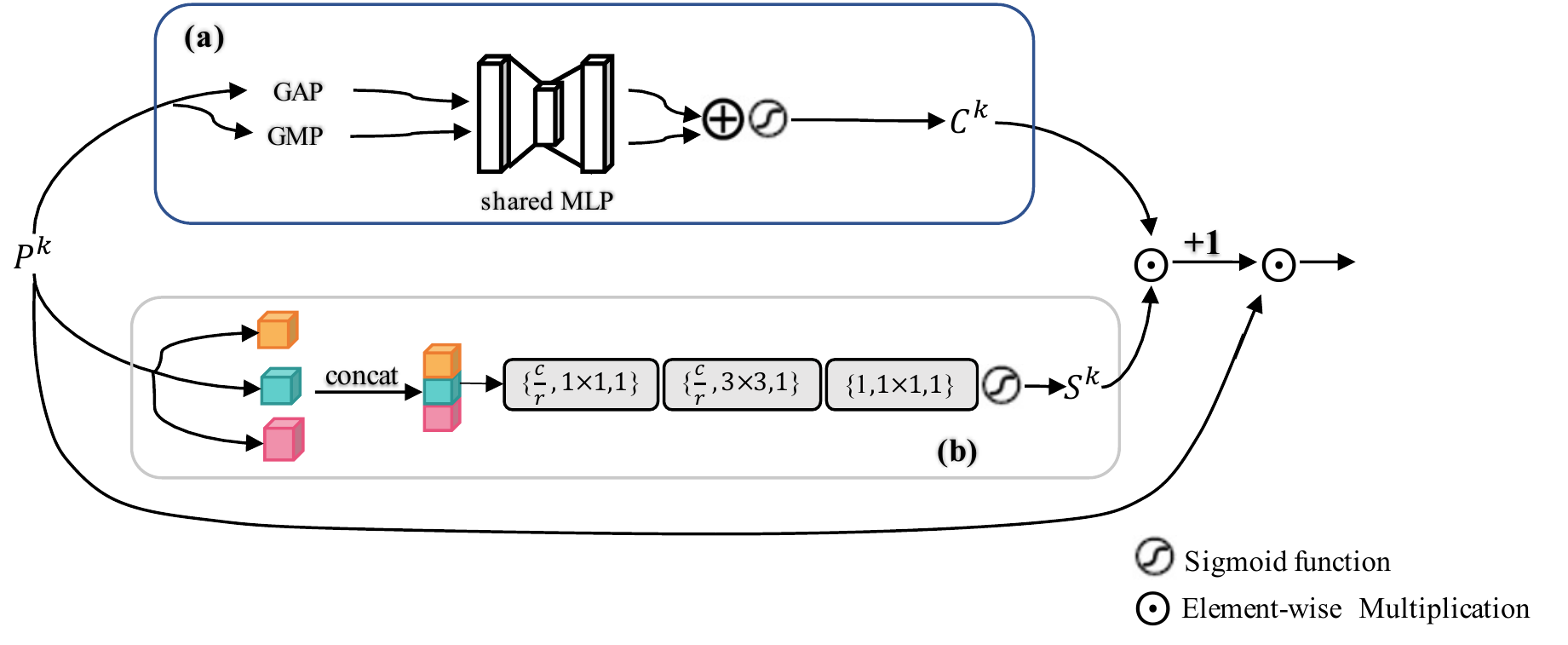}
\caption{Structure of Multi-scale Attention Module. (a) Similar to CBAM\cite{cbam}, the channel block utilizes both Global Max-Pooling (GMP) and Global Average-Pooling (GAP) outputs with a shared Multi-Layer Perceptron (MLP) to get the channel mask, $C^k$. (b) In the spatial block, $P^k$ is first fed into three $3\times3$ Conv layers with different dilation ratios (1,2 and 3) to mine multi-scale features. Then, the output goes through three Conv layers and a Sigmoid function to obtain the spatial mask, $S^k$. The three items in the bracket of a Conv layer are filter number, filter shape, and stride, respectively. The ReLU and Batch Normalization applied to each Conv layer are not shown for brevity.}
\label{figMAM}
\end{figure}

\textbf{Multi-scale Attention Module. } Current Re-ID methods \cite{r26,rattn2} 
ignore the fact that vehicle components often exhibit multiple scales or various shapes, and small visual regions might be easily missed, especially in CNN's top layers. To relieve this, we adopt a MAM in Student and Teacher branches of each stream. Based on CBAM \cite{cbam}, MAM learns attention masks in a joint but factorized way. The overall formulation can be computed as
\begin{equation}
   \mathbf{MAM}(P^k)=(C^k \odot S^k+1) \odot P^k,
\end{equation}
where $C^k \in \mathbb{R}^{1\times 1 \times c}$ denotes the channel attention mask, $S^k \in \mathbb{R}^{h\times w \times 1}$ indicates the spatial attention mask, and $\odot$ is the element-wise multiplication with repeating data in $h$ and $w$ dimension for $C^k$ and $c$ dimension for $S^k$. As is illustrated in Fig.\ref{figMAM}, the channel attention block is the same as the one in CBAM \cite{cbam}. In the spatial attention block, to make region scales uniform, we first feed $P^k$ into three $3\times3$ Conv layers with multiple dilation ratios (1, 2, and 3) and concatenate the features. The concatenated features containing multi-scale information are fed into another three Conv layers and a sigmoid function to obtain $S^k$.


\subsubsection{Teacher-Student Guided Learning via Part Transfer Loss}
\label{SecPartTransfer}
In each part-level stream of PMNet, Teacher branches guide their Student branches by distilling the vehicle part concept into Students. The observation is that although the location of a certain vehicle part varies between images, the mean position in a mini-batch remains stable. In other words, the average location of one vehicle part is consistent across batches. Hence, we propose a batch-wise Part Transfer loss to constrain the similarity of the feature maps between Teacher and Student branches. In particular, in the $k$-th stream, we first apply a global average operation over the batch and channel dimension of the output features from Students, $\Tilde{F}_i^k$, and Teachers, $\Tilde{P}_i^k$. The formula can be written as,
\begin{small}
\begin{equation}
\Tilde{F}^{k}=\frac{1}{C}\sum_{1}^{C}\left(\frac{1}{B}\sum_{1}^{B}\Tilde{F}_i^k\right), 
\Tilde{P}^{k}=\frac{1}{C}\sum_{1}^{C}\left(\frac{1}{B}\sum_{1}^{B}\Tilde{P}_i^k\right),
\end{equation}
\end{small}where $C$ and $B$ is the number of channel and the batch size, respectively.

Afterwards, we apply our Part Transfer loss on the pixel level to constrain the similarity such that it transfers the partial knowledge. This process can be computed as
\begin{small}
\begin{equation}
\mathbf{\mathcal{L}_{PT}}=\frac{1}{H}\sum_{i=1}^{H}\left(\frac{1}{W} \sum_{i=1}^{W}\left\|\Tilde{P}^{k}-\Tilde{F}^{k}\right\|_{2}\right).
\label{eq:PTL}
\end{equation}
\end{small}

This loss encourages Students and Teachers to pull each other closer such that Teachers can transfer specific vehicle part information to Students.

\subsection{Multi-task Learning for PMNet}
\label{sec3C}
Regarding vehicle Re-ID, PMNet produces three types of features: global feature $f_G$, the concatenated feature $f_{S}$ from Students, and the concatenated feature $f_{T}$ from Teachers ($f_G$, $f_S$ and $f_T$ are marked in Fig.\ref{Fig:PGLN}). We propose multi-task learning for these features. All of them are supervised by the Cross-Entropy loss ($\mathcal{L}$) and Triplet loss ($\mathcal{L}_{Tri}$) \cite{r46}, as shown in Fig. \ref{Fig:PGLN}. Obviously, it is essential to balance the weighting of each ID loss. Instead of using a naïve linear sum of multiple objective losses or labor-intensive grid search, we leverage Homoscedastic Uncertainty Learning (HUL) \cite{r6} to automatically learn the optimal weights. We first obtain the output Softmax probability $\hat y^j, j\in\{G,S,T\}$ by applying three FC layers to features $f_G$, $f_S$, and $f_T$, respectively. Then, HUL introduces a noise parameter $\sigma_j$ for each $\hat y^j$. As the noise $\sigma_j$ increases, the loss weight $\alpha_j$ for its respective objective decreases. Hence, the minimization objective of the HUL-based multi-task ID loss can be converted to

\begin{equation}
\mathbf{\mathcal{J}_{ID}}\approx\sum_{j\in\{G, S, T\}}\frac{1}{\sigma_{j}^{2}} \mathbf{\mathcal{L}^{j}}\left(\hat{y}^{j},y\right)+\sum_{j\in\{G, S, T\}}\operatorname{\log}\sigma_{j},
\end{equation}where $\hat{y}^{j}$ is the Softmax probability for the sub-task $j$, $y$ is the identity one-hot label for one sample, and $j\in\{G,S,T\}$ indicates the ID loss for global feature learning, Students' part feature learning, and Teachers' part feature learning, respectively. Similarly, Triplet loss for the $j$-th sub-task can be calculated as,

\begin{equation}
\mathbf{\mathcal{J}_{Tri}}=\frac{1}{3}(\mathbf{\mathcal{L}_{Tri}}(f_G)+\mathbf{\mathcal{L}_{Tri}}(f_S)+\mathbf{\mathcal{L}_{Tri}}(f_T)).
\end{equation}

In total, the multi-task ID loss, Triplet loss \cite{r46}, and Part Transfer loss (Eq. (\ref{eq:PTL})) are combined to facilitate training of PMNet. The overall objective $\mathbf{\mathcal{J}}$ can be formulated as
\begin{equation}
\begin{aligned}
\mathbf{\mathcal{J}}=
&\mathbf{\mathcal{J}_{ID}}+\mathbf{\mathcal{J}_{Tri}}+\mathbf{\mathcal{L}_{PT}}.
\label{Eqt11}
\end{aligned}
\end{equation}

\section{Experiments}
\label{SecEXP}

In this section, we first introduce two widely used vehicle Re-ID datasets and evaluation metrics. Then, we detail the implementation of the training and inference of our method. Finally, we compare our method with the state-of-the-arts and conduct ablation studies to validate the effectiveness of each component.

\subsection{Datasets and Evaluation Metrics}
\subsubsection{Datasets}
We use two datasets to evaluate the performance of our method.

\textbf{VeRi776} \cite{r27} consists of around 51,035 bounding boxes of 776 vehicles, including 576 identities in the training set and 200 in the test set. The standard probe and gallery set contain 1,678 and 11,579 images, respectively. Images in this dataset are collected by 20 cameras across block regions under different viewpoints, making it one of the most popular vehicle Re-ID datasets. Additionally, it provides other vehicle information, including model, color, and trajectory clues.

\textbf{VehicleID} \cite{r50} contains totally 221,763 images for about 26,267 vehicles. The test set is divided into three sizes (small, medium, and large). During inference, for each vehicle identity, one image is randomly selected as the gallery image while the others are regarded as query set. Images of this dataset are either captured under front or rear view.

\textbf{Occluded vehicle testset} contains a small testset designed by \cite{r55} and a larger one which we construct by applying random color patches to the query set of VeRi776 as shown in Fig.\ref{fig:occluded}. The small test set totally includes 60 identities and 600 images, with 300 for query and 300 for gallery, whilst the larger one contains 200 identities and 11579 images, with 1678 for query and the rest of the images for gallery.

\subsubsection{Evaluation Metrics}

We evaluate models' performance with two types of Re-ID metrics: mean Average Precision (mAP) and Cumulative Matching Characteristics (CMC). In specific, mAP calculates the averaged area under the Precision-Recall (AP) curve for all the query images. CMC@1 and @5 measure the probability of locating at least one true positive in the top-1 and top-5 ranks, respectively.

\subsection{Implementation Details}
\subsubsection{Training}
Implemented with PyTorch, our model is based on the well-known Bag-of-Tricks Re-ID baseline\footnote{\url{https://github.com/DTennant/reid_baseline_with_syncbn}} \cite{r43}, named Baseline in our following experiments. We trained PANet for 100 epochs on VeRi776. The batch size is 64 and the learning rate is $1\times10^{-4}$. In PMNet, we randomly sampled $P$ identities and $Q$ images per vehicle to constitute a training batch. Finally, the batch size $B = P\times Q$. Here, we set $P = 4$, $Q = 8$, so$B = 32$. The base learning rate of PMNet is $1.5\times10^{-4}$ with a warm-up strategy, and images are resized to $256\times256$. For data augmentation, we apply random erasing augmentation and random horizontal flip with a probability of $0.5$. As regards other hyper-parameters in our experiments, the number of vehicle parts localized by PANet, $K$, is set to 3; the margin for Triplet loss \cite{r46}, $\beta$, is set to 0.7. In addition, we use Adam as the optimizer.

Since there is no special dataset for occluded vehicle Re-ID, we simulate the occlusion by using a simple data augmentation technique with a probability of $0.3$. For each training batch, we first randomly select an area as the obstruction region, and these areas are then filled with random color patches.

\subsubsection{Inference}
During inference, we obtain $f_G$ for global feature and the concatenated features $f_{S}$ and $f_{T}$ for part features, respectively. The final distance between a probe and a gallery image can be calculated as \(\lambda_{1} D_{f_G}+\lambda_{2} D_{f_S}+\lambda_{3} D_{f_T}\), where $\lambda_{i}$, $i\in \{1, 2, 3\}$ denotes the optimal weights learned by HUL and $D(.)$ indicates the Cosine Distance of the features between two images. On VeRi776, $\lambda_1=1.81$, $\lambda_2=2.21$, $\lambda_3=2.25$; On VehicleID, $\lambda_1=
5.06$, $\lambda_2=4.59$, $\lambda_3=4.62$. Our method with ``PMNet only'' in Table.\ref{table1} and Table.\ref{table2} means that PANet and the Teacher branches in PMNet are removed, i.e., $\lambda_{3}=0$. All the experiments are conducted on a platform with 256GB RAM and 2$\times$Intel Xeon Silver 4214R CPU @ 2.40GHz. The GPU we use is a single GeForce RTX 2080Ti with 11GB VRAM.


\subsection{Comparison with State-of-the-art Methods}
\label{SecExpVeri776}
We first compare our PMANet with a variety of recent approaches on VeRi776 \cite{r27} and VehicleID \cite{r50}.

        
        

        


\subsubsection{Experiments on VeRi776}
We adopt CMC@1, CMC@5 and mAP as the evaluation protocol on VeRi776. Table.\ref{table1} outlines the annotations used by the state-of-the-art methods. Accordingly, we divide the methods into two categories in terms of whether using extra annotations. PRReID\cite{r9}, SPAN\cite{r10}, MVAN\cite{r52}, CFVMNet\cite{r53}, PVEN\cite{r11}, and TBE-Net\cite{r75} rely on extra annotations, such as viewpoints, vehicle part bounding boxes, or expensive keypoints. RAM\cite{r8}, MRM\cite{r77}, SAN\cite{r12}, SAVER\cite{r54}, HPGN\cite{r74}, 
Baseline\cite{r43} and our PANet+PMNet are trained without extra manual labels. It is notable that CFVMNet, RAM, SAN, and DDM also incorporate other vehicle attributes (e.g., vehicle model, color) to enhance feature expressions. 

\begin{table}\footnotesize
\caption{Performance (\%) comparisons with state-of-the-art methods on the VeRi776 \cite{r27} benchmark. ``View'': viewpoint annotations, ``Attri.'': attributes, and ``RR'': re-ranking technique \cite{r28}.}
\begin{center}
  \centering
   \setlength{\tabcolsep}{1.0mm}{
    \begin{tabular}{l|l|ccc|c}
    \hline
        Method & Annotations & mAP & CMC@1 & CMC@5 & Year \\ \hline\hline

        PRReID\cite{r9}  &ID+Part Boxes & 70.2 & 92.2 & 97.9& 2019 \\
        SPAN\cite{r10}  &ID+View & 68.9 & 94.0 & 97.6& 2020 \\ 
        MVAN\cite{r52}   &ID+View & 72.5 & 92.6 & 97.9& 2020 \\
        CFVMNet\cite{r53}  &ID+View+Attri. & 77.1 & 95.3 & 98.4& 2020 \\
        PVEN\cite{r11}  &ID+Keypoints &79.5 & 95.6 & 98.4& 2020 \\
        TBE-Net\cite{r75} &ID+Part Boxes & 79.5 &96.0 & 98.5 &2021\\\hline
        
        RAM\cite{r8}   & ID+Attri. & 61.5 & 88.6 & 94& 2018 \\
        MRM\cite{r77} &ID & 68.6 &91.8 &95.8 &2019\\
        SAN\cite{r12}   &ID+Attri. & 72.5 & 93.3 & 97.1& 2020 \\
        SAVER\cite{r54}  & ID & 79.6 & 96.4 & 98.6& 2020\\
        HPGN\cite{r74} &ID & 80.2 & \textbf{96.7} & - & 2021 \\\hline
        
        Baseline\cite{r43}  &ID & 77.2 & 95.7 & 97.9& 2019\\
        
        PMNet only  &ID & 81.5 & 96.6 & \textbf{98.6} & Ours \\ 
        PANet+PMNet   &ID & \textbf{81.6} & 96.5 & \textbf{98.6} & Ours \\\hline \hline
        CFVMNet\cite{r53}+RR  &ID+View+Attri. & 81.5 & 94.8 & 96.6& 2020 \\
        PMNet+RR   & ID & \textbf{89.4} & \textbf{97.0} & \textbf{98.3} & Ours \\ \hline 
    \end{tabular}}
\end{center}

\label{table1}
\end{table}

From the table, we can observe that our PANet+PMNet outperforms state-of-the-art methods without extra annotations in mAP, e.g., our method surpasses Baseline and HPGN by 4.4\% and 1.4\% in mAP, respectively. The significant improvement from Baseline is caused by the proposed attention-based vehicle part localization and teacher-student guided learning. The CMC@1 of our method is comparable with that of HPGN. Besides, by removing Teacher branches during inference, only PMNet yields an even higher CMC@1. Additionally, we adopt a post-processing re-ranking (RR) technique \cite{r28}, denoted as PMNet+RR (see the last entry of the table). As a result, the mAP significantly rises to 89.4\%, and the CMC@1 increases to 97.0\%. 

\begin{figure}[t]
\centering
\includegraphics[scale=0.7]{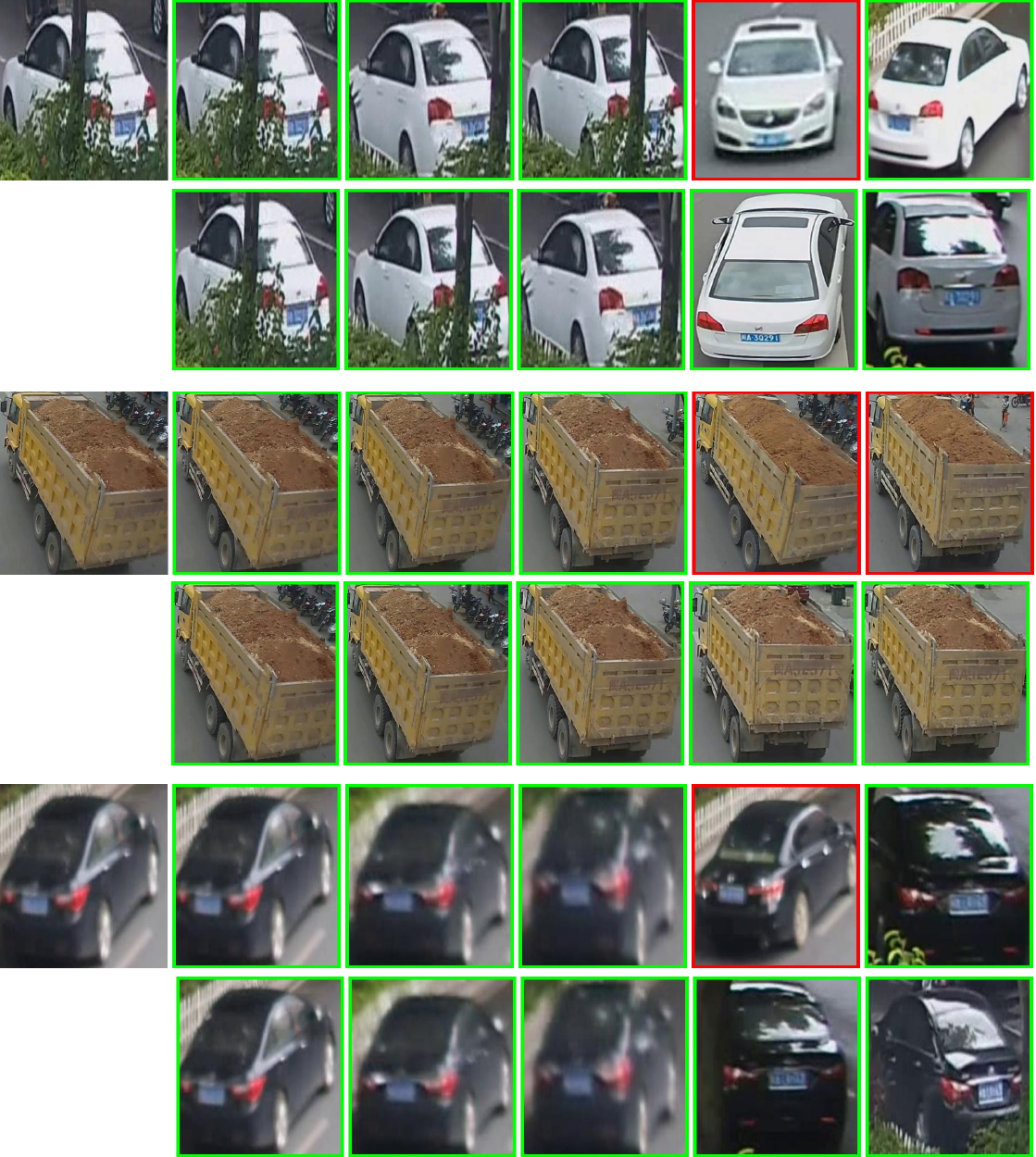}
\caption{Visualization of vehicle Re-ID ranking list on VeRi776 \cite{r27}. The images on the first column are the query images and the rest show retrieved gallery images. For each query sample, the first and second row respectively shows the top-5 results by Baseline \cite{r43} and our PMNet. The correct and false matched vehicle images are enclosed in \textcolor{green}{green} and \textcolor{red}{red} rectangles, respectively. }
\label{figVisualqg}
\end{figure}

Fig.\ref{figVisualqg} demonstrates some sample vehicle Re-ID results on VeRi776. Compared with the retrieved ranking list of the baseline, it is clear that our method produces more reliable results. These top-5 results show that our method is more robust to viewpoint variation, low resolution, and background clutter and is also more capable of mining fine-grained local clues to distinguish near-identical vehicles.

\begin{table}\footnotesize
\caption{Performance (\%) comparisons with state-of-the-art methods on VehicleID \cite{r50}. View indicates viewpoint annotations while Attri. is short for attributes. The evaluation metric is CMC.}
\begin{center}
   \centering
   \setlength{\tabcolsep}{1.2mm}{
    \begin{tabular}{l|l|cc|cc|cc}
    \hline
        \multirow{2}*{Method} &
        \multirow{2}*{Annotations} &
        \multicolumn{2}{c|}{Small} &
        \multicolumn{2}{c|}{Medium} &
        \multicolumn{2}{c}{Large} \\
        \cline{3-8}
        & & @1 & @5 & @1 & @5 & @1 & @5 \\ \hline\hline

        TAMR\cite{r26} & ID+Central points & 66.0 & 79.7 & 62.9 & 76.8 & 59.7 & 73.9 \\
        MVAN\cite{r52} & ID+View & - & - & -& - &72.6 & 83.1 \\
        
        PRReID\cite{r9} & ID+Part Boxes & 78.4 & 92.6 & 75.0 & 88.3 & 74.2 & 86.4 \\
        CFVMNet\cite{r53} & ID+View+Attri. & 81.4 & 94.1 & 77.3 & 90.4 & 74.7 & 88.7 \\
        ATT\cite{r83} & ID+Part Boxes & 79.7 & - &77.5 & - & 74.8 & -\\
        PVEN\cite{r11} & ID+Keypoints & 84.7 & 97.0 & 80.6 & \textbf{94.5} & \textbf{77.8} & 92.0\\\hline

        RAM\cite{r8} & ID+Attri. & 75.2 & 91.5 & 72.3 & 87.0 & 67.7 & 84.5 \\
        MRM\cite{r77} &ID & 76.6 &92.3 &74.2 &88.5 &70.9 &84.8 \\
        SAN\cite{r12} & ID+Attri. & 79.7 & 94.3 & 78.4 & 91.3 & 75.6 & 88.3 \\
        SAVER\cite{r54} & ID & 79.9 & 95.2 & 77.6 & 91.1 & 75.3 & 88.3 \\
        HPGN\cite{r74} &ID &83.9 &- &80.0 & - & 77.3 & -\\\hline
        
        PMNet only & ID & 85.2 & \textbf{97.5} & \textbf{80.7} & \textbf{94.5} & 77.7 & 91.8 \\
        PANet+PMNet & ID & \textbf{85.3} & 97.3 & 80.5 & \textbf{94.5} & 77.6 & \textbf{92.2} \\ \hline
    \end{tabular}}
\end{center}

\label{table2}
\end{table}

\subsubsection{Experiments on VehicleID}
Since there is only one ground truth for each query vehicle in VehicleID, only CMC@1 and CMC@5 are compared in this dataset. Table.\ref{table2} illustrates the comparison results on the small, medium, and large test sets. Table.\ref{table2} outlines the annotations required by recent methods and groups them into two categories according to the usage of extra manual labels. Results show that our proposed PMNet beats all the models requiring no additional annotations, with an average improvement of 0.8\% in CMC@1 and 3.0\% in CMC@5. Among methods using extra annotations (e.g., central points, vehicle part boxes, keypoints, and viewpoints), our method performs best on the small test set with an increase of 0.6\% in CMC@1 and 0.5\% in CMC@5. As for the medium and large test sets, our method produces comparable performance to PVEN, even though our method does not require dense key-point labels that PVEN does. Such a weakly-supervised manner makes our approach more suitable for practical applications. It is also notable that our proposed ``PMNet only'' can bypass vehicle part localization (i.e, PANet) during inference, thus saving computational cost.

\subsection{Ablation Study}
\subsubsection{Component Analysis}
\label{SecComponent}
We conduct comparative experiments to validate the effectiveness of the proposed components, including PANet, and the global feature learning, part feature learning and MAM in PMNet. Detailed results are tabulated in Table.\ref{table3}. 

\begin{table}[t]\footnotesize
\caption{Ablation studies about the key components of our PANet and PMNet on VeRi776 \cite{r27}. $\checkmark$ in each row denotes that the corresponding component is included in this experiment. $\mathcal{R}_1$ indicates that PANet is replaced by Uniform Division for part mask generation. $\mathcal{R}_2$ denotes that all Student branches in PMNet are removed during both training and inference. Exp-7 denotes our entire method (PANet+PMNet).}
\begin{center}
   \centering
   \setlength{\tabcolsep}{0.5mm}{
   \rowcolors{2}{gray!10}{white}
    \begin{tabular}{c|cccc|ccc}
    \hline
        \footnotesize{\makecell[c]{Experiment\\Number}}
        & \footnotesize{PANet} 
        & \footnotesize{\makecell[c]{Global feature\\learning}} 
        
        & \footnotesize{\makecell[c]{Part feature\\learning}} 
        & \footnotesize{MAM} 
        & \footnotesize{mAP} & \footnotesize{CMC@1} & \footnotesize{CMC@5} \\ \hline\hline
        Exp-1 &  &$\checkmark$ &  &  &72.4 & 94.6 & 97.2  \\
        Exp-2 & $\checkmark$ & $\checkmark$ &$\mathcal{R}_2$  &  &78.6 & 95.5 & 98.0  \\ 
        Exp-3 & $\checkmark$ & $\checkmark$ & $\checkmark$ &  &79.5 & 96.4 & 98.2 \\ 
        Exp-4 & $\mathcal{R}_1$ & $\checkmark$  & $\checkmark$ & $\checkmark$ & 79.9 & 96.3 & 98.3  \\ 
        Exp-5 & $\checkmark$  &  & $\checkmark$ & $\checkmark$ & 78.1 & 95.3 & 97.8  \\ 
        Exp-6 & $\checkmark$ & $\checkmark$ & $\mathcal{R}_2$ & $\checkmark$ & 78.3 & 95.7 & 98.1  \\ 
        Exp-7 & $\checkmark$ & $\checkmark$ & $\checkmark$ & $\checkmark$ & \textbf{81.6} & \textbf{96.5} & \textbf{98.6}  \\ \hline
    \end{tabular}}
\end{center}
\label{table3}
\end{table}

\begin{figure}[t]
\centering
\subfigure[]{
    \label{fig7-a} 
    \includegraphics[scale=0.39]{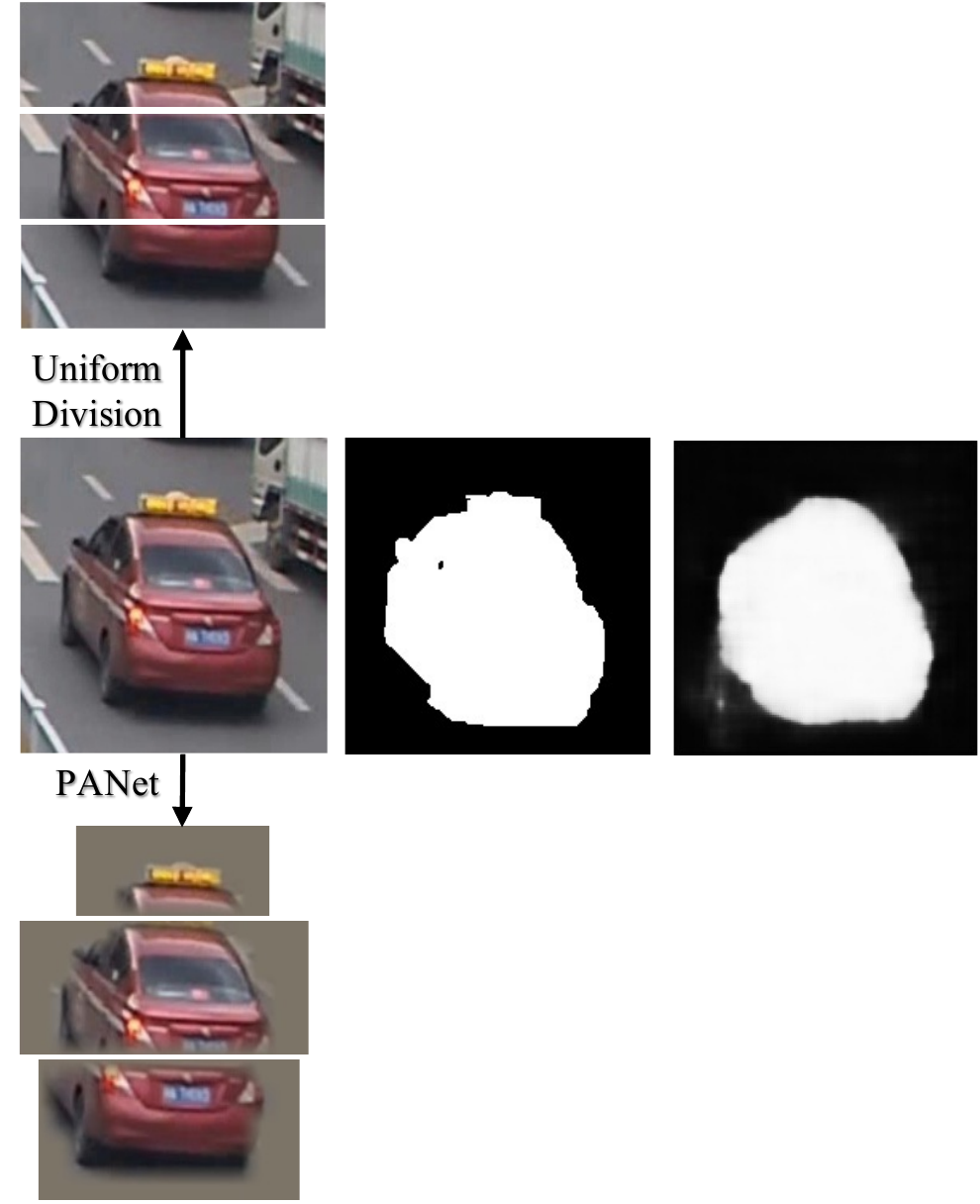}
    }\quad
  \subfigure[]{
    \label{fig7-b} 
    \includegraphics[scale=0.39]{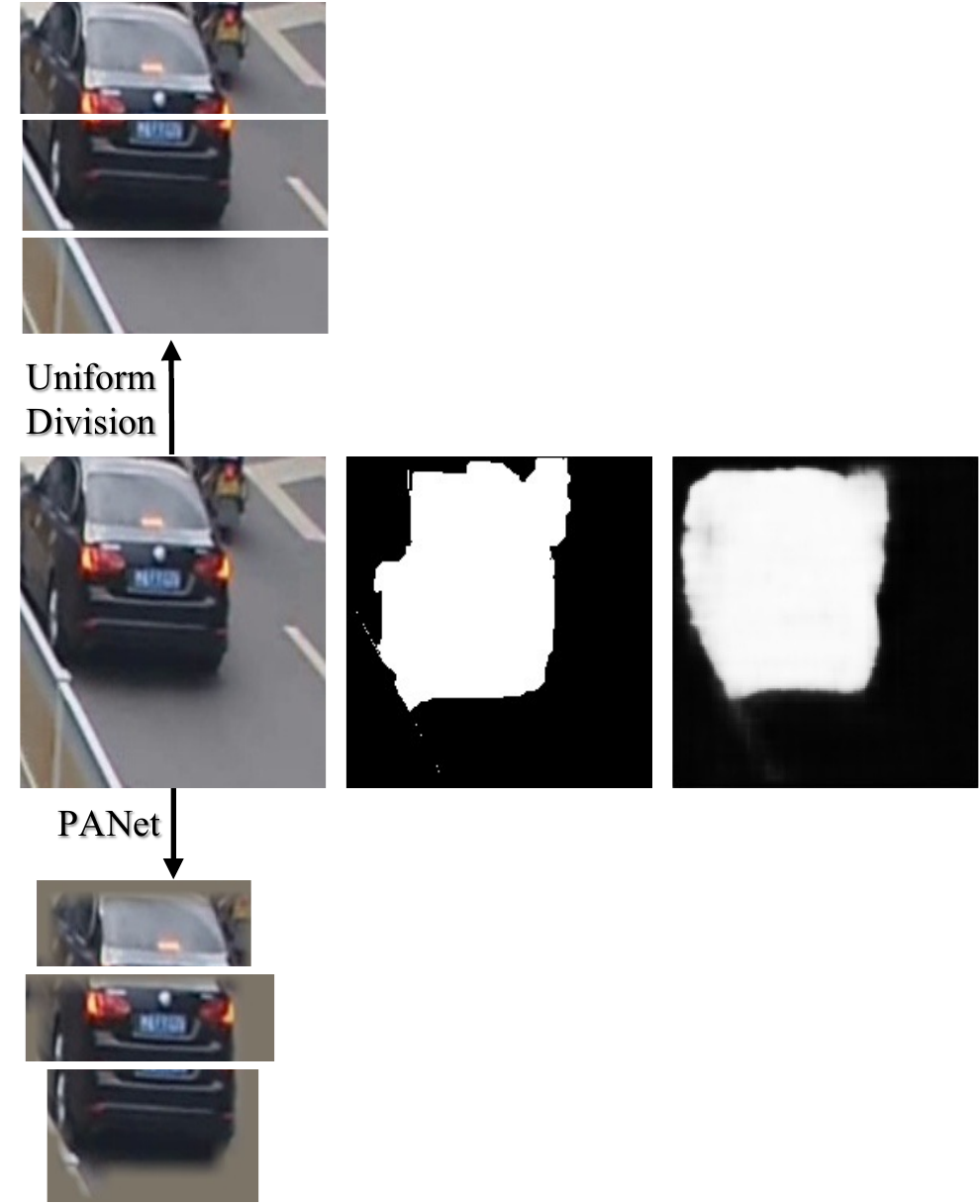}
    }

\caption{Two examples of generated masks by Uniform Division \cite{r8,r12} and our PANet. In (a) and (b), figures on the first row illustrate three vehicle parts vertically split by Uniform Division. The three images on the second row respectively represent the sample image, the coarse foreground mask by GrabCut \cite{r40} and the refined one by our PANet. The figures on the third row exhibit the $K$ part masks predicted by PANet. $K=3$ in this experiment.}
\label{fig7}
\end{figure}

\textbf{Part Attention Network.} PANet produces three part masks for PMNet to perform teacher-student style guided learning. We replace PANet with a simple Uniform Division \cite{r8,r12}, namely $\mathcal{R}_1$, which evenly splits the feature map into three stripes, as mask generation in Exp-4. As is shown in Table.\ref{table3}, Exp-7 (our entire method) beats Exp-4 by 1.7\% in mAP and 0.2\% in CMC@5. Moreover, such a naive splitting strategy is unstable. As the first row of Fig.\ref{fig7-a} and Fig.\ref{fig7-b} depict, especially when a car body is unevenly distributed in the image (see Fig.\ref{fig7-b}), Uniform Division suffers from spatial misalignment and might miss some crucial information. From the third row, in contrast, our PANet is able to locate different salient vehicle parts (e.g., vehicle roof, windscreen, lights) with almost all the subtle cues, such as personalized decorations and inspection marks. With the help of the refined foreground masks, our three dense part masks are robust to background clutter and noises.

\textbf{Global feature learning head in PMNet.} From Exp-5 and Exp-7, we can observe a dramatic increase of 3.5\%, 1.2\% and 0.8\% in mAP, CMC@1 and CMC@5, respectively. These results validate the effectiveness of the extracted global feature in PMNet, showing that global and part-level features facilitate each other by providing complementary information.

\textbf{Part feature learning head in PMNet.} Instead of using part teacher-student guided learning, Exp-2 and Exp-6 remove Student branches and retain Teachers branches in the case with and without MAM for local feature extraction. Without MAM, comparing Exp-2 with Exp-3, our proposed teacher-student guided learning leads to an increase of 0.9\% in mAP and 0.9\% in CMC@1. Comparing Exp-6 to Exp-7 with MAM, our proposed teacher-student manner boosts the performance by 3.3\% in mAP and 0.8\% in CMC@1. These two comparisons validate that our part feature learning manner has a superior capability of extracting part-level clues than plain single (Teachers) branches.

\begin{figure}[t]
\centering
\subfigure[]{
    \label{fig8-a} 
    \includegraphics[scale=0.38]{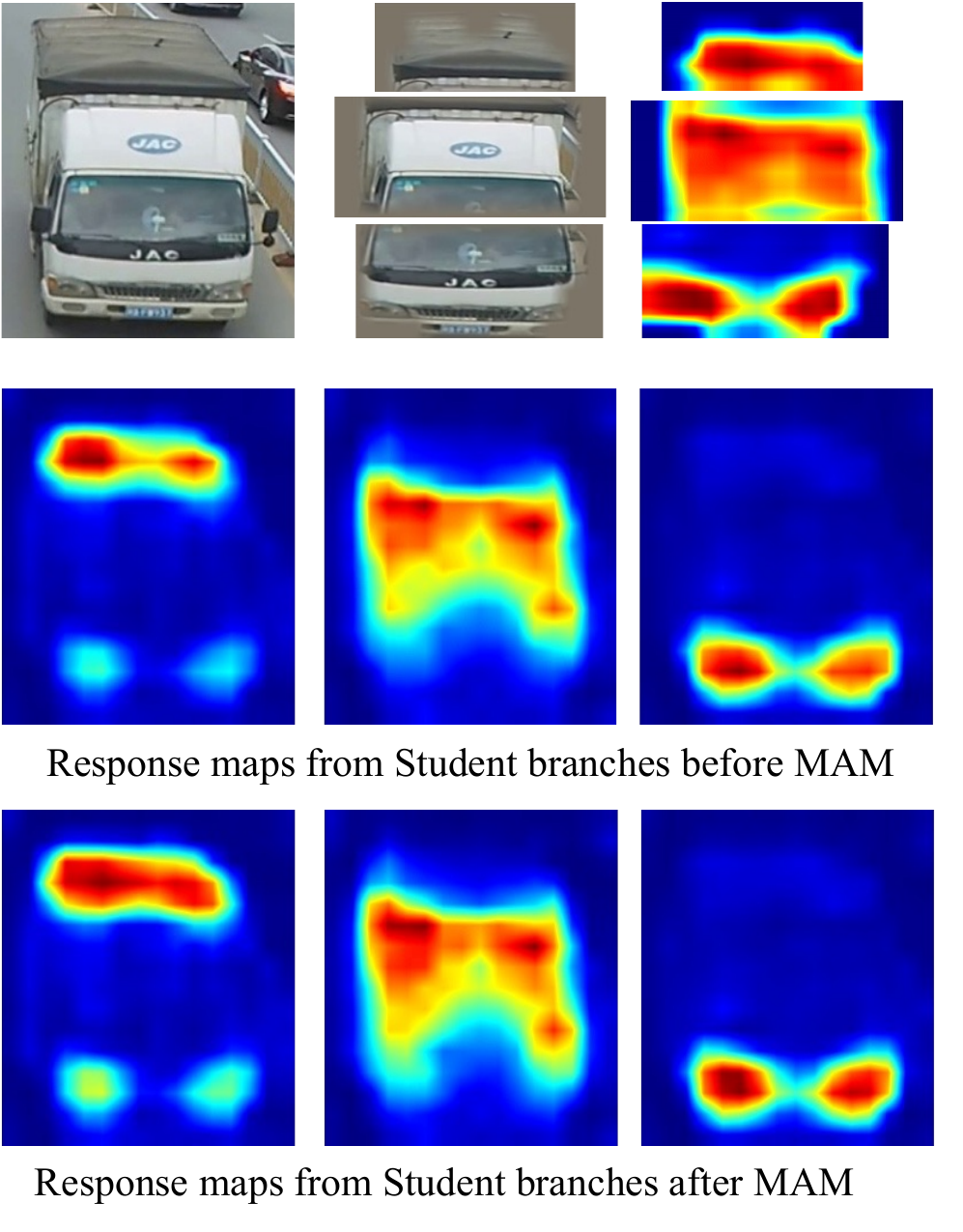}
    }\quad
  \subfigure[]{
    \label{fig8-b} 
    \includegraphics[scale=0.38]{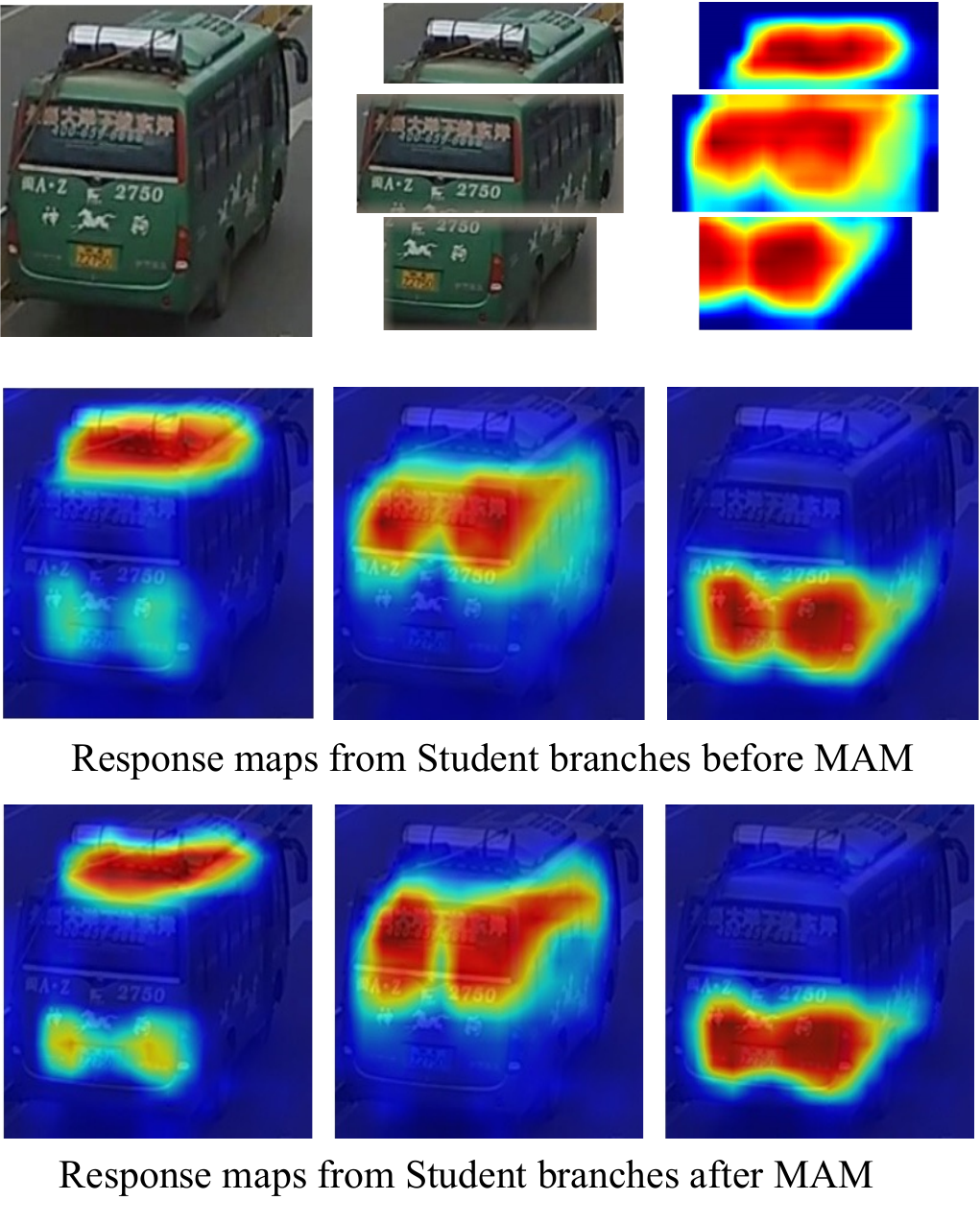}
    }

\caption{Two examples of visualized attention maps. In (a) and (b), figures on the first row show the original image, three part masks predicted by PANet, and three attention maps from Teacher branches after MAM. The second and the third rows show three Student branches' attention maps before and after MAM, respectively.}
\label{fig8}
\end{figure}

\textbf{Multi-scale Attention Module.} 
Apart from the part-level attention in PANet, MAM in each Teacher and Student branch of PMNet performs a second closer look at these vehicle parts for pixel-wise local features to decrease inter-class similarity especially between near-identical images. Comparing Exp-7 with Exp-3, MAM brings an improvement of 2.1\% in mAP. Additionally, we visualize attention maps on three streams of PMNet to explore how our MAM affects part feature learning, and the visualized maps are shown in Fig.\ref{fig8}. By comparing attention maps before and after the MAM in each Student branch, we can observe that MAM assists in capturing and amplifying more subtle clues in localized vehicle parts, like the roof, windscreen and car-lights in Fig.\ref{fig8-a}.

\begin{table}\footnotesize
\caption{Comparison experiments to validate Multi-scale Attention Module (MAM) on VeRi776 \cite{r27}. For the first and second entries, MAM is replaced by SE-Net \cite{r23} and Residual Attention Module adapted in TAMR \cite{r26}, respectively.}
\begin{center}
   \centering
   \setlength{\tabcolsep}{1.5mm}{
    \begin{tabular}{l|ccc}
    \hline
        Method & mAP \\ \hline \hline
        PMNet w/ SE-Net \cite{r23} & 81.0 \\
        PMNet w/ Residual Attention Module \cite{r26} & 80.7  \\
        PMNet w/ MAM (ours) & \textbf{81.6} \\ 
 \hline
    \end{tabular}}
\end{center}

\label{tableAttn}
\end{table}

Moreover, we compare MAM with other state-of-the-art attention methods to verify its superiority. Among existing approaches in vehicle Re-ID, SE-Net \cite{r23} is a channel-wise attention commonly utilized in vision models, and TAMR \cite{r26} leverages Residual Attention Module as pixel-wise refinement in its regional feature learning. We replace MAM with these two attention modules, as shown in the first and second entries of Table.\ref{tableAttn}. Accordingly, the mAP score drops by 0.6\% and 0.9\%. Therefore, compared with SE-Net and  Residual Attention Module \cite{r26}, our MAM exhibits better ability in handling complex scale variation and mining multi-grained clues.

\subsubsection{Superiority of Part Transfer Loss}
Unlike learning features with plain Conv layers in a single branch \cite{r26,r10,r11}, our PMNet introduces three teacher-student pairs that leverage parameter sharing and the Part Transfer loss to convey vehicle part-relevant information from Teachers to Students. Part information transfer aims to align the feature space of each teacher-student pair such that Students can learn to focus on the specified vehicle parts instead of the holistic features. 

\begin{table}[t]\footnotesize
\caption{Effectiveness of parameter sharing and our Part Transfer loss, $\mathbf{\mathcal{L}_{PT}}$, on VeRi776 \cite{r27}. \emph{w/o} denotes without.}
\begin{center}
   \centering
   \setlength{\tabcolsep}{1.5mm}{
    \begin{tabular}{c|ccc}
    \hline
        Method & mAP & CMC@1 & CMC@5 \\ \hline \hline
         w/o parameter sharing \& $\mathbf{\mathcal{L}_{PT}}$  & 80.6 &96.4 & 98.3 \\
        w/o parameter sharing & 81.1 &96.3 & 98.4 \\
        w/o $\mathbf{\mathcal{L}_{PT}}$ & 80.2  & \textbf{96.7}  & \textbf{98.6} \\
        PANet+PMNet & \textbf{81.6} & 96.5 & \textbf{98.6} \\ 
 \hline
    \end{tabular}}
\end{center}
\label{table4}
\end{table}

From Fig.\ref{fig8}, the attention maps of each teacher-student pair mainly concentrate on the same vehicle part, and the three streams concentrate on three different regions located by PANet. This means that the parameter sharing and the novel Part Transfer loss successfully guides the learning of Students with concept conveyed by Teachers. Besides, we conduct experiments to verify the effectiveness of parameter sharing and Part Transfer loss, $\mathbf{\mathcal{L}_{PT}}$, in Table \ref{table4}. With a respective increase of 1.4\%, 0.5\% in mAP, $\mathbf{\mathcal{L}_{PT}}$ and parameter sharing prove to be beneficial.

         


\begin{table}[t]\footnotesize
\caption{Experiments to investigate multi-task learning with Homoscedastic Uncertainty Learning on VeRi776 \cite{r27}. \emph{WR} is short for weighting ratio. }
\begin{center}
   \centering
   \setlength{\tabcolsep}{1.5mm}{
   \rowcolors{2}{white}{gray!10}
    \begin{tabular}{c|ccc}
    \hline
        Method & mAP & CMC@1 & CMC@5 \\ [3pt]\hline\hline
        \makecell[c]{w/o HUL\\(\emph{WR}:1:1:1:1:1)} & 80.5 & 96.5 & 98.0 \\[3pt]
        \makecell[c]{w/o HUL\\(\emph{WR}:2:1:1:1:1)} & 80.4 & 96.2 & \textbf{98.7} \\ [3pt]
        \makecell[c]{w/o HUL\\(\emph{WR}:4:1:1:1:1)} & 79.8 & 96.4 & 98.0 \\[3pt]
        \makecell[c]{Ours with HUL} & \textbf{81.6} & \textbf{96.5} & 98.6 \\[3pt] \hline
    \end{tabular}}
\end{center}

\label{table6}
\end{table}

\subsubsection{Multi-task Learning with Homoscedastic Uncertainty Learning} We model this Re-ID issue as three sub-tasks, i.e., global feature learning, Students' part feature learning, and Teachers' part feature learning. With the shared backbone, these three different tasks can be trained end-to-end in our unified feature learning network, PMNet, and achieve the optimal model generalization with the help of Homoscedastic Uncertainty Learning (HUL) \cite{r6}. Here we investigate the multi-task learning with HUL in Table.\ref{table6} on VeRi776. Specifically, we remove HUL and manually select three different weights for the first three rows in Table.\ref{table6}. Obviously, usage of HUL yields an over 1\% increase in mAP. Apart from speeding the convergence process during training, HUL helps save the time cost by extra manual tuning and is robust to hyper-parameter changes.

\begin{figure}[t]
\centering
\subfigure[]{
    \label{fig:occluded-a} 
    \includegraphics[scale=0.40]{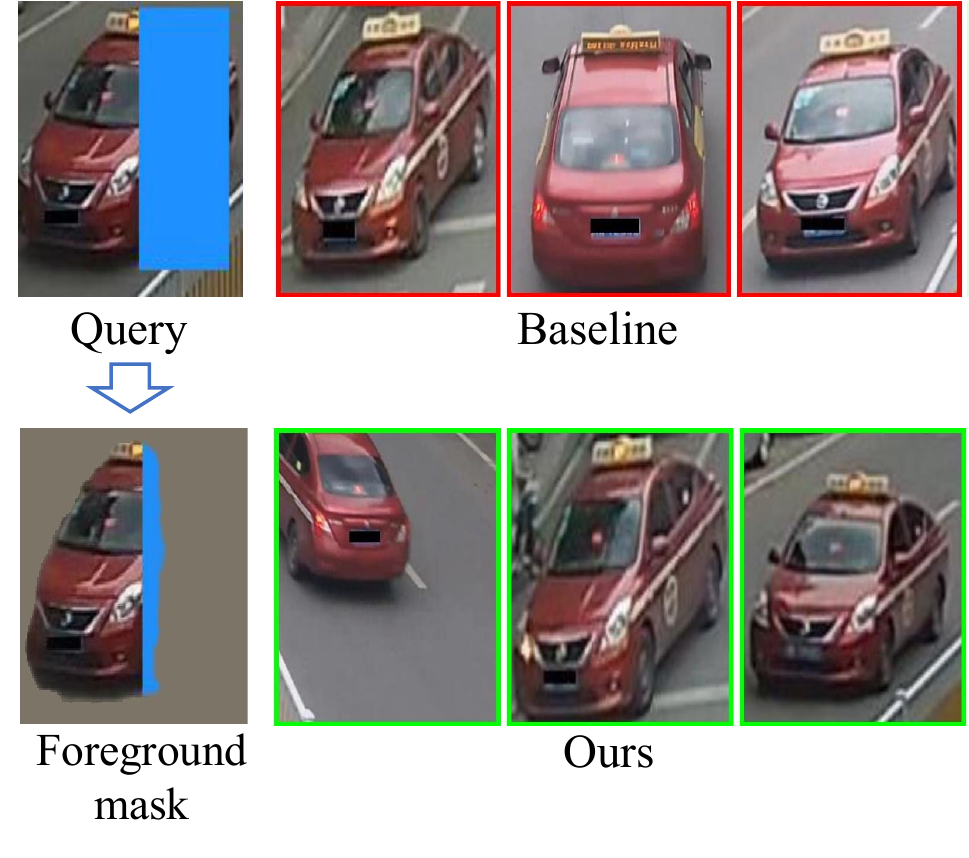}
    }\quad
  \subfigure[]{
    \label{fig:occluded-b} 
    \includegraphics[scale=0.40]{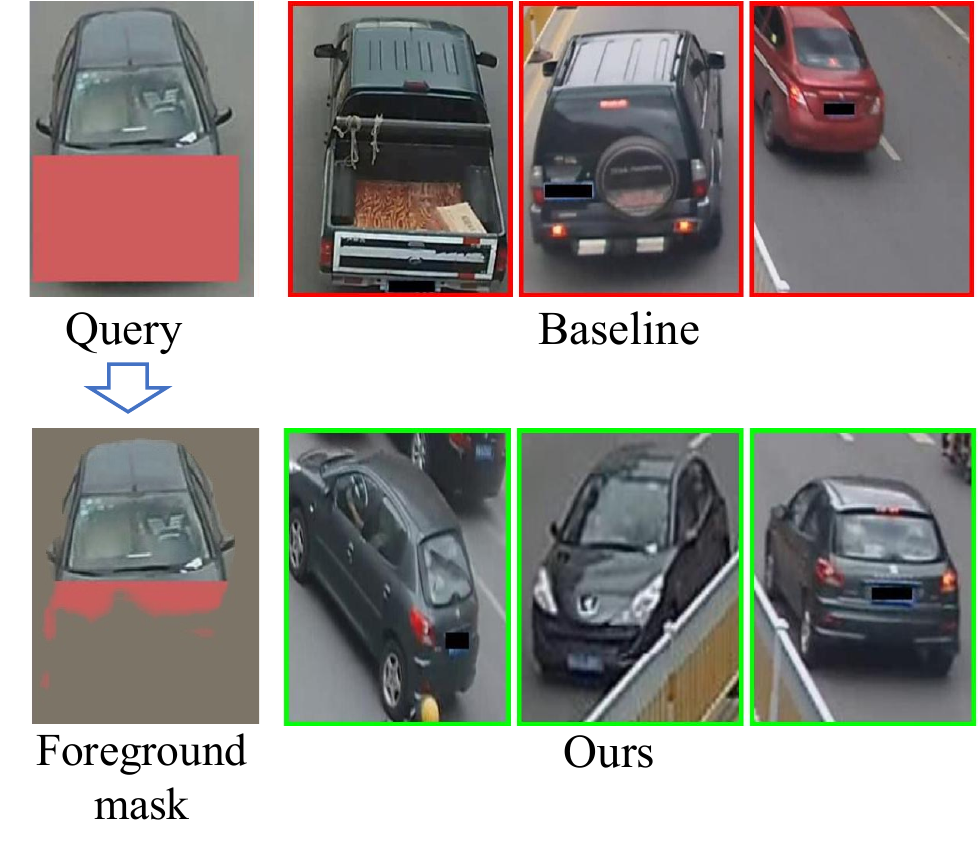}
    }
\caption{Visualization results on two occluded vehicle images on \cite{r55}. In (a) and (b), images on the first column exhibit the query and the foreground mask generated by PANet. Images on the second column show the top-3 retrieval results of the Baseline \cite{r43} and our method. \textcolor{green}{Green} and \textcolor{red}{red} rectangles indicate correct and wrong results, respectively.}
\label{fig:occluded}
\end{figure}

\begin{table}[t]\footnotesize
\caption{Comparison experiments in occluded vehicle Re-ID. For the small test set, we use the same test set designed by \cite{r55}. With 1678 identities for query, the large test set is constructed based on the test set of VeRi776 \cite{r27}}
\begin{center}
   \centering
   \setlength{\tabcolsep}{1.3mm}{
    \begin{tabular}{c|ccc|ccc}
    \hline
        \multirow{2}*{Method} &
        \multicolumn{3}{c|}{Small} &
        \multicolumn{3}{c}{Large} \\
        \cline{2-7} 
        & mAP & CMC@1 & CMC@5 & mAP & CMC@1 & CMC@5 \\ \hline \hline
        ASAN \cite{r55} & 53.3 & 68.8 & 90.1 &-&-&- \\
        Baseline \cite{r43}  & 60.1  & 70.7  & 90.7 & 54.5 &75.9 & 85.8 \\
        PANet+PMNet & 85.3 & \textbf{96.0} & \textbf{99.7} & 65.7 & 85.3 & 91.6 \\ 
        PMNet only & \textbf{85.7} & \textbf{96.0} & \textbf{99.7} & \textbf{66.2} & \textbf{86.0} & \textbf{91.8} \\ 
 \hline
    \end{tabular}}
\end{center}

\label{tableOccluded}
\end{table}

\subsection{Occluded Vehicle Re-ID}
\label{Sec-occluded}
To validate the robustness towards occlusion, we firstly show some visualization results of the Baseline \cite{r43} and our proposed method in Fig.\ref{fig:occluded}. We can observe that the Baseline fails on two test occluded images mainly because it regards the occlusion color as the texture of the query image. The foreground masks generated by PANet successfully avoid the occlusion color blocks. Therefore, our method effectively achieves robustness to occluded scenarios. To give a fair comparison, we compare our method with the Baseline and the state-of-art ASAN \cite{r55} on the occluded small testset designed by \cite{r55} and the large testset, respectively. As is tabulated in Table \ref{tableOccluded}, our ``PMNet only'' and ``PANet+PMNet'' surpass the others with a significant improvement of over 25\% in both mAP and CMC@1 on the small testset. On the large testset, our method performs best with an increase of over 10\% in mAP, verifying the generalization ability of our model.

\section{Conclusion}
\label{SecConclusion}
In this paper, we propose a weakly-supervised Part Attention Network (PANet) for vehicle part localization and a Part-Mentored Network (PMNet) for teacher-student guided feature learning and aggregation. PANet locates informative vehicle parts under weak supervision through part-relevant channel recalibration and cluster-based mask generation. To address the two weaknesses of using plain convolutional branches for part feature learning, PMNet builds one Student branch and Teacher branch for each vehicle part. Teachers guide their Students so that PMNet can bypass prior vehicle part prediction during inference. This Re-ID issue is modeled as three sub-tasks under a shared backbone. Experimental results demonstrate that our approach outperforms state-of-the-art methods. 
Moreover, results on the occluded vehicle Re-ID test set show the robustness to background interference.

\ifCLASSOPTIONcaptionsoff
  \newpage
\fi



\bibliographystyle{IEEEtran}

\bibliography{IEEEabrv,main}

\end{document}